\documentclass[10pt,twocolumn,letterpaper]{article}

\usepackage{cvpr}      % To produce the REVIEW version
\usepackage{algorithm}
\usepackage{algorithmic}
\usepackage{lipsum} % 用于生成占位文本，删除
\usepackage{xcolor} % ，删除
\definecolor{lightgray}{gray}{0.95} % 0.9表示非常浅的灰色，删除
\usepackage{booktabs}
\usepackage{multirow}
\usepackage{amsmath,amssymb} 
\usepackage{xspace}
\makeatletter
\DeclareRobustCommand\onedot{\futurelet\@let@token\@onedot}
\def\@onedot{\ifx\@let@token.\else.\null\fi\xspace}
\def\eg{\emph{e.g}\onedot} 
\def\ie{\emph{i.e}\onedot}

\def\etal{\emph{et al}\onedot}
\makeatother
\usepackage{array}

\definecolor{cvprblue}{rgb}{0.21,0.49,0.74}
\usepackage[pagebackref,breaklinks,colorlinks,allcolors=cvprblue]{hyperref}

\def\paperID{12554} % *** Enter the Paper ID here
\def\confName{CVPR}
\def\confYear{2025}

\title{RDPM: Solve Diffusion Probabilistic Models via Recurrent Token Prediction}

\author{Xiaoping Wu$^1$\;\;\;\;\;\; Jie Hu$^{2,3,1}$\thanks{Corresponding author and project lead.}\;\;\;\;\; Xiaoming Wei$^1$\\
$^1$Meituan \;\; $^2$Institute of Software, Chinese Academy of Sciences \;\; $^3$UCAS\\
{\tt\small \{wuxiaoping03, weixiaoming\}@meituan.com\;\; hujie@ios.ac.cn}
}

\begin{document}
\maketitle

% \begin{center}
% \begin{figure*}[t]
% \includegraphics[width=1\linewidth]{./tmp.pdf}
% \caption{
% Visualizations of the $256 \times 256$ images generated by RDPM.
% }
% \end{figure*}
% \end{center}
% \begin{center*}
% \centering
% \includegraphics[width=1\linewidth]{./tmp.pdf}
% \caption{
% Visualizations of the $256 \times 256$ images generated by RDPM
% }
% \end{center*}

% \begin{strip*}
% % \begin{figure*}[t]
% \centering
% \includegraphics[width=1\linewidth]{./tmp.pdf}
% \captionof{figure}{Feature graphic caption.
% \label{fig:feature-graphic}}
% % \end{figure*}
% \end{strip*}

\begin{abstract}
%
% Diffusion probabilistic models (DPMs) play a crucial role in the success of high-quality image synthesis.
% However, numerous necessary sampling steps are time-consuming and computationally expensive.
%
% In this work, we propose a novel generation paradigm called RDPM that simulates the diffusion process via recurrent token prediction within limited $10$ time-steps.
%
% During the tokenization phase, the RDPM progressively adds Gaussian noise to the image and represents it with vector-quantized tokens.
% %
% In the generative part, starting from pure noise, a transformer model learns to iteratively predict the token codes of next data distribution.
% %
% Extensive experiments on the ImageNet dataset for class-conditional image generation task demonstrate the superiority of our method.
% %
% The code and model weights will be public to the open-source community.
% %
%
Diffusion Probabilistic Models (DPMs) have emerged as the \textit{de facto} approach for high-fidelity image synthesis, operating diffusion processes on continuous VAE latent, which significantly differ from the text generation methods employed by Large Language Models (LLMs). In this paper, we introduce a novel generative framework, the Recurrent Diffusion Probabilistic Model (RDPM), which enhances the diffusion process through a recurrent token prediction mechanism, thereby pioneering the field of \textit{Discrete Diffusion}. By progressively introducing Gaussian noise into the latent representations of images and encoding them into vector-quantized tokens in a recurrent manner, RDPM facilitates a unique diffusion process on discrete-value domains. This process iteratively predicts the token codes for subsequent timesteps, transforming the initial standard Gaussian noise into the source data distribution, aligning with GPT-style models in terms of the loss function. RDPM demonstrates superior performance while benefiting from the speed advantage of requiring only a few inference steps.  This model not only leverages the diffusion process to ensure high-quality generation but also converts continuous signals into a series of high-fidelity discrete tokens, thereby maintaining a unified optimization strategy with other discrete tokens, such as text. We anticipate that this work will contribute to the development of a unified model for multimodal generation, specifically by integrating continuous signal domains such as images, videos, and audio with text. We will release the code and model weights to the open-source community.
\end{abstract}
% --------------------------------------
\begin{figure}[!t]
    \centering
    \includegraphics[trim=5.5cm 6.2cm 6.5cm 3.2cm, clip, width=0.98\linewidth, keepaspectratio]{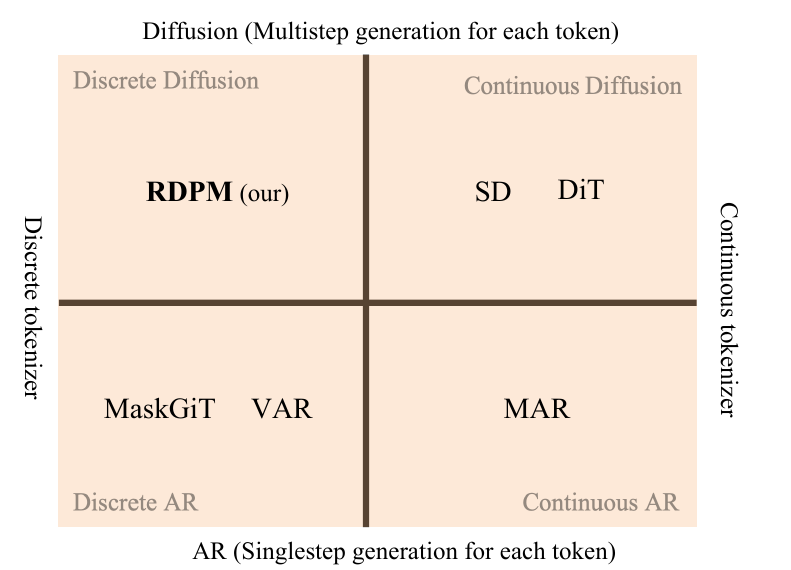}
    \caption{
        Taxonomy of modern image generation. From the continuous/discrete tokenizer and AR/Diffusion perspective, we categorize modern generation approaches into four types, including continuous diffusion, discrete diffusion, continuous AR, and discrete AR.
    }
\label{fig:taxonomy}
\vspace{-10pt}
\end{figure}
% --------------------------------------

% --------------------------------------
\begin{figure}[!t]
\centering
\includegraphics[width=0.98\linewidth, keepaspectratio]{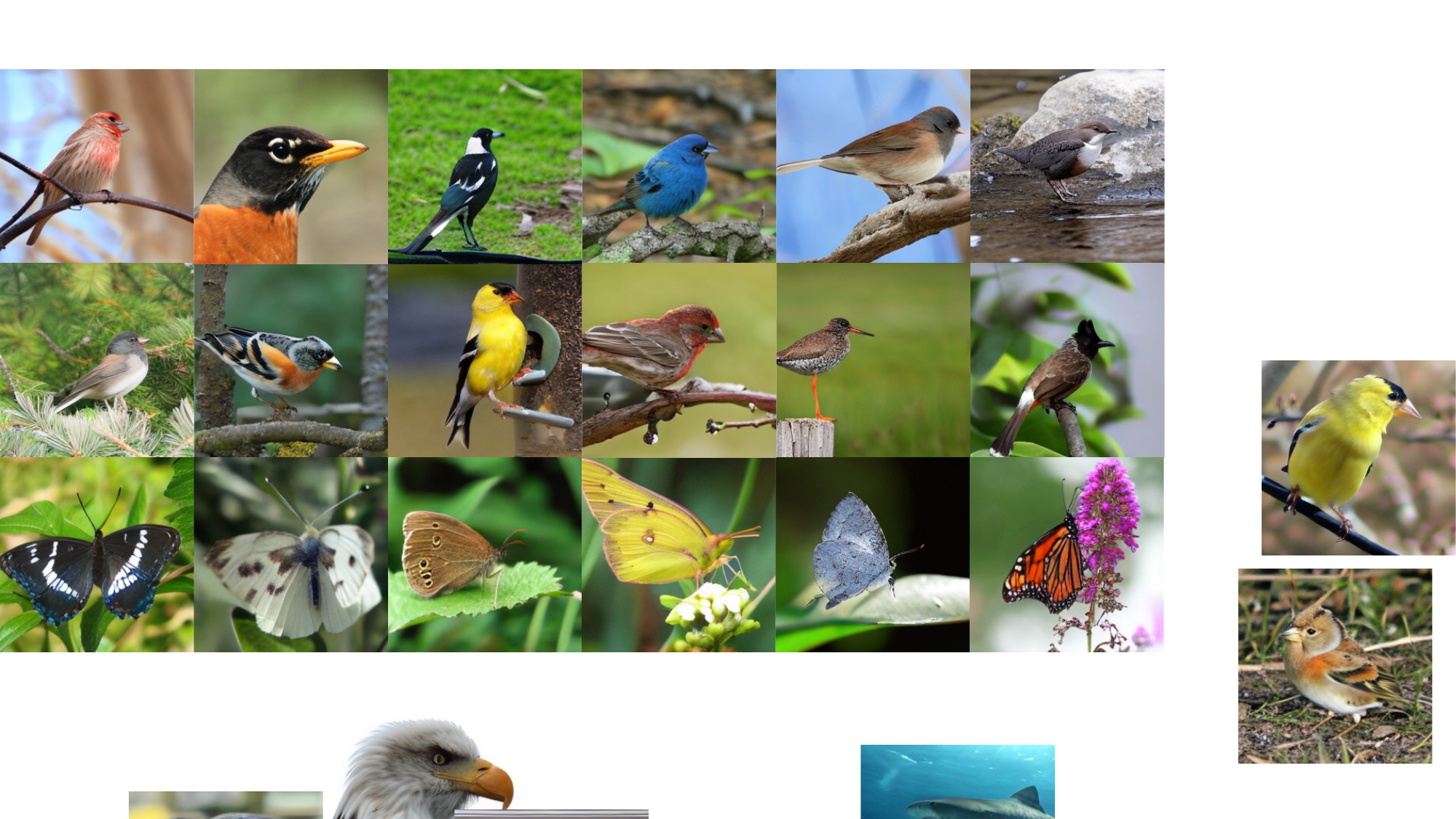}
\caption{
Visualizations of the ImageNet class-conditional $256 \times 256$ images generated by RDPM.
}
\label{fig:vis_res}
\vspace{-10pt}
\end{figure}
% --------------------------------------

% --------------------------------------
\begin{figure*}[!t]
\centering
\includegraphics[width=1.0\linewidth, keepaspectratio]{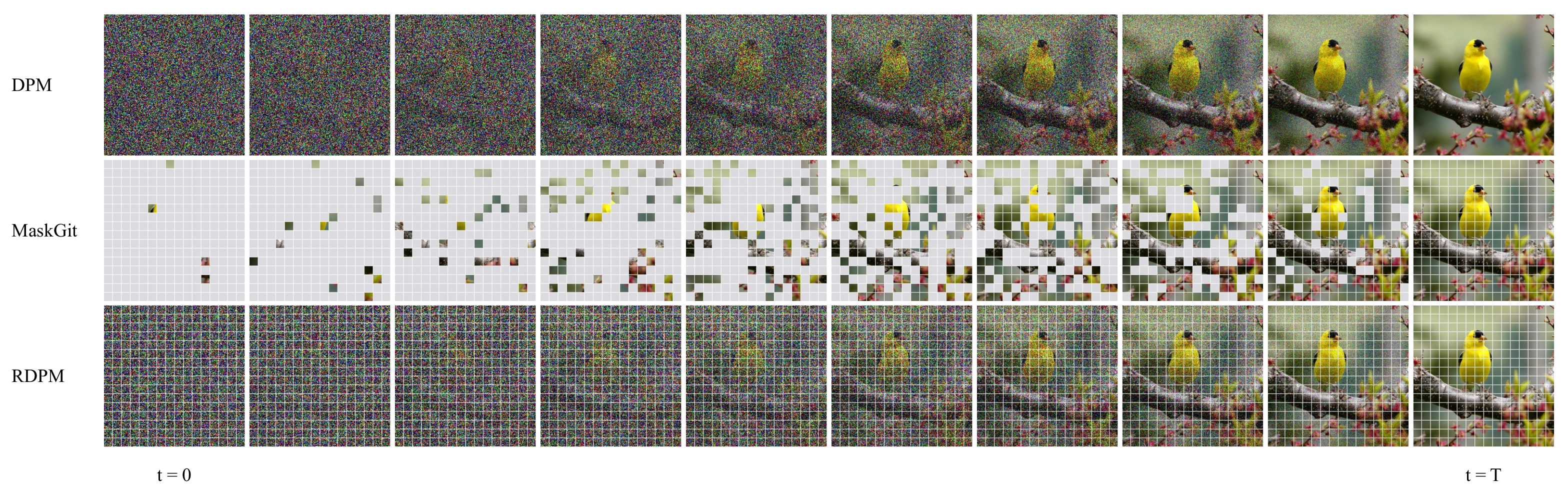}
\caption{
Comparison of various visual generation patterns. The images in rows 2 and 3, segmented into blocks, represent discrete quantized tokens, whereas the images in row 1 are depicted with raw pixels. The gray grids denote masked tokens.
}
\label{fig:cmp}
\vspace{-10pt}
\end{figure*}
% --------------------------------------

\vspace{-10pt}
\section{Introduction}

The advent and evolution of diffusion models~\cite{ho2020denoising,ho2022classifier,ho2022cascaded,rombach2022high} have significantly advanced the generation of high-fidelity and high-resolution images. To further enhance performance, DiT~\cite{peebles2023scalable} introduced a pure transformer architecture to replace the traditional UNet structure, facilitating the scalability of subsequent models such as SD3.0~\cite{esser2024scaling} and FLUX~\footnote{https://huggingface.co/black-forest-labs/FLUX.1-dev}. Concurrently, efficient sampling methods~\cite{song2020denoising,song2020score,nichol2021improved,lu2022dpm} have been proposed to reduce the number of diffusion steps required.

To achieve structural compatibility with large language models (LLMs)~\cite{touvron2023llama,jiang2023mistral}, autoregressive (AR) methods~\cite{tschannen2023givt,yu2023language,esser2021taming,yu2021vector} employ vector quantization (VQ)~\cite{van2017neural,razavi2019generating} during the encoding phase. These methods infer one timestep per image token, in contrast to the gradual denoising required by diffusion processes. Directly applying an autoregressive strategy on raster scanning sequences is inherently time-consuming. For example, a $256\times256$ resolution image, resolution image, typically represented by $16\times16$ tokens, necessitates 256 inferences, with this number escalating dramatically as image resolution increases.
To unlock the potential of transformer-based generative models in real-time or resource-constrained environments, MaskGit~\cite{chang2022maskgit} employs a parallel token decoding scheme. As illustrated in Fig.~\ref{fig:cmp}, the model divides the image synthesis process into a fixed number of steps, independently predicting vector quantization codes for random sub-patches at each step while masking others. This approach significantly enhances image generation efficiency.
Subsequently, VAR~\cite{tian2024visual} introduced another image generation strategy based on AR, using a multi-scale autoregressive paradigm with next-scale prediction, akin to a GPT-style AR model. This strategy achieves remarkable results through a coarse-to-fine multistep prediction strategy in resolution. We refer to this approach, along with MaskGit, which performs AR in discrete-valued domains, as \textbf{Discrete AR}. A common trait of these methods is that they execute only one prediction per token at the output, resulting in lower image generation fidelity compared to conventional diffusion models. Conventional diffusion models utilizing VAE as an image tokenizer are classified as \textbf{Continuous Diffusion}.

A pertinent question arises: \textit{Is it possible to combine the advantages of diffusion methods and AR methods?} MAR~\cite{li2024autoregressive} exemplifies work in this area, modeling per-token probability distribution on continuous-valued domains through a subsequent small diffusion process module, thereby pioneering a new field with significant results. However, despite performing AR on continuous-valued domains, its loss remains diffusion-based, differing from the cross-entropy loss used in LLMs. This discrepancy is not optimal for developing unified models for multimodal generation. We categorize this approach of performing AR on continuous-valued domains, despite a minimal diffusion module, as \textbf{Continuous AR}. Fig.~\ref{fig:taxonomy} illustrates the taxonomy of modern image generation methods.

In this work, we explore a previously uncharted area—\textbf{Discrete Diffusion}—and propose a novel image generation paradigm termed RDPM (Recurrent Diffusion Probabilistic Model). Specifically, we progressively introduce noise in latent space during image tokenization until it becomes pure noise. We then employ a Recurrent VQ-VAE to model this process, where the representation of discrete features in multistep residuals performs comparably to the continuous representation of VAE. Additionally, the multistep discrete tokens can ensure that the subsequent diffusion process can utilize cross-entropy loss, aligning with the GPT-style approach. During inference, starting from standard Gaussian noise, we execute the denoising process of the diffusion model. Unlike vanilla diffusion with continuous values, each step predicts discrete codes, gradually restoring the natural data distribution. On the ImageNet $256\times256$ benchmark, RDPM achieves superior performance compared to other models using discrete image tokenizers. Fig.~\ref{fig:vis_res} shows qualitative results. However, we emphasize that the primary objective of this work is not to achieve absolute state-of-the-art results across all domains of image generation, but to propose a new paradigm for image generation. We outline three principal contributions of our work:

\begin{itemize} 
\item We propose a novel method for solving diffusion probabilistic models that diverges from vanilla denoising in continuous space. Instead, we perform denoising in discrete space, thereby preserving the same loss function form as the autoregressive (AR) model. This approach provides new insights for developing a unified model that integrates both continuous and discrete signal generation.

\item Within RDPM, we introduce an innovative image discretization method—diffusion-based image tokenization—that effectively mitigates information loss associated with the discretization of continuous features. This method is applicable to various continuous signal domains, including video and audio.

\item Our method achieves state-of-the-art results among approaches utilizing discrete visual encoders. Additionally, it delivers quality comparable to conventional diffusion models while significantly enhancing computational efficiency by compressing the diffusion process into just 10 timesteps.
\end{itemize}
\section{Related Work}

Image synthesis~\cite{ho2020denoising,song2020denoising} has become a pivotal area of research in the pursuit of general artificial intelligence, aiming to generate high-quality and realistic visual content. This section reviews three primary approaches in image generation: diffusion, autoregression, and mask-based mechanisms.

\subsection{Diffusion-based Image Synthesis}

Diffusion models operate in two phases: a forward process where noise is incrementally added to an image, simulating diffusion, and a reverse process that restores image details from noise through denoising. Early models like GANs~\cite{goodfellow2020generative,arjovsky2017wasserstein,brock2018large,kang2023scaling,sauerscaling} and VAEs~\cite{kingma2013auto} directly learned data distributions. In contrast, DDPMs~\cite{ho2020denoising} utilize parameterized Markov chains for data generation, achieving high image quality while avoiding training instability and mode collapse. However, DDPMs lag in log-likelihood performance compared to likelihood-based models.

Subsequent improvements~\cite{nichol2021improved} in inverse diffusion variance, hybrid learning objectives, and noise schedules enhanced DDPM's coverage of target distributions. DDIM~\cite{song2020denoising} further accelerated sampling by generalizing DDPM to a non-Markov chain diffusion scheme, optimizing variance, and reducing the need for randomness, thus enabling deterministic generation with fewer denoising steps. The Consistency Model~\cite{song2023consistency} and Latent Consistency Model (LCM)~\cite{luo2023latent} introduced methods to reduce sampling steps without sacrificing quality, utilizing classifier-free guidance and LoRA~\cite{hu2021lora} for efficient training.

In conditional image synthesis, Dhariwal and Nichol~\cite{dhariwal2021diffusion} employed classifiers to guide diffusion models, while CFG~\cite{ho2022classifier} proposed end-to-end training for conditional guidance. Traditional models like U-Net~\cite{ronneberger2015u} faced scalability challenges, leading to the adoption of transformer architectures~\cite{peebles2023scalable} in models such as SD3~\cite{esser2024scaling}, Pixart~\cite{chen2023pixart}, and Kolor. Diffusion models excel in high-resolution image generation, balancing quality and efficiency.

\subsection{AutoRegression-based Image Synthesis}

Autoregressive (AR) models, successful in LLMs~\cite{brown2020language,achiam2023gpt,touvron2023llama}, are increasingly applied to image synthesis. VQ-VAE~\cite{van2017neural} introduced encoding images into discrete tokens using a codebook for vector quantization. VQ-VAE2~\cite{razavi2019generating} refined this with hierarchical reconstruction, while VQ-GAN~\cite{esser2021taming} integrated perceptual loss and adversarial training. Yu~\etal~\cite{yu2021vector} further enhanced this by employing vision transformers and improving loss functions.

Despite their strengths, AR models face challenges with information loss during quantization, addressed by RQ-VAE~\cite{lee2022autoregressive} through residual quantization. AR models are adept at capturing dependencies in sequence data and integrating with LLMs for multi-modal models, but face challenges with parallelization and embedding conditions affecting efficiency and quality.

\subsection{Mask-based Image Synthesis}

Mask-based generative models, like MaskGit~\cite{chang2022maskgit}, represent a novel approach in image synthesis, predicting image positions in parallel but with varying validity. During training, random masking allows the model to infer unknown regions from known patches. Inference begins with reliable patches, gradually completing the image. This approach is utilized in models like MAGVIT~\cite{yu2023magvit} and Muse~\cite{chang2023muse}. Li~\etal~\cite{li2023mage} demonstrated that joint training of generative and representation learning models enhances performance.

While mask-based models improve efficiency and resource utilization, they lack the iterative refinement of diffusion models and the long dependency capture of AR models. Inspired by these methods, our proposed RDPM innovatively models the diffusion process into a recurrent token prediction framework, requiring only 10 iterations. Through parallel forward passes, RDPM progressively converges from pure noise to the data distribution.

\section{Methodology}
Our image generation approach comprises two key components: a diffusion-based image tokenization module that encodes images into discrete tokens, and a recurrent token prediction module that serves as a diffusion mechanism for image generation.

\vspace{5pt}
\subsection{Diffusion-based Image Tokenization}
Building on existing research~\cite{ho2020denoising,song2020denoising}, our method performs image generation in the latent space rather than the pixel space, significantly reducing data dimensionality and computational resource consumption. The initial step involves training a VAE for image encoding and decoding.

\vspace{5pt}
\noindent\textbf{Image Encoding}:
The encoder $E$ compresses natural images into feature representations, mapping an input image $x \in \mathbb{R}^{H \times W \times 3}$ to a continuous latent vector $v \in \mathbb{R}^{h \times w \times d}$, where $H/h$ is the downsampling ratio, typically 16, compressing a $256 \times 256$ image to a $16 \times 16$ spatial dimension.
\begin{equation}
v = E(x).
\end{equation}
The encoder $E$ is primarily composed of residual blocks interleaved with downsampling modules. Additionally, we incorporate layer normalization at the end of the encoder module to ensure that the information within the latents can be disrupted by the Gaussian noise introduced during the subsequent diffusion process.

\vspace{5pt}
\noindent\textbf{Diffusion-based Quantization Process}:
This subsection aims to divide the original single-step vector quantization operation into a  $T$-step diffusion process. Instead of adding noise directly to the original image, we apply the diffusion paradigm within the quantization process. This strategy minimally increases the parameter volume and computational cost of the original VAE model. Specifically, we mix Gaussian noise into the latent variable $v$ in $T$ steps.

At each step $t$, let $v_t$ and $v'_t$ denote the noisy vector and its corresponding quantized vector, respectively:
\begin{equation}
v_t = \alpha_t z_t + \beta_t \epsilon_t,
v'_t = Q(v_t).
\end{equation}
Here,  $z_t=z_{t-1}-v'_{t-1}$ performs a residual operation similar to RQVAE~\cite{lee2022autoregressive}, with $z_1=z$. The quantization operation 
 $Q(\cdot)$ searches for the nearest neighbor in the codebook $C$ (containing $K$ embeddings) based on the Euclidean distance, resulting in a code $C_t$ of size $h\times w \times 1$. Additionally, $\epsilon_t \sim \mathcal{N}(0, 1)$ is noise sampled from a standard normal distribution. $\alpha_t$ is a weight controlling the image proportion, which increases from small to large values, and when $t = T$, $\alpha_t = 1$, with $\beta_t^2 + \alpha_t^2 = 1$.
That is, when $t$ is small, the main component of the noisy vector is Gaussian noise, and as it approaches $T$, the image content increases. During the $T$ steps of repetition, the model continuously accumulates operations on $v'_t$, \ie, $z'_t = \sum_{i=1}^{t} v'_i$. This represents the current step's image quantization representation.
Similar to RQVAE~\cite{lee2022autoregressive}, the purpose of the residual and accumulation operations on the latent vector is to minimize the loss of information during the quantization process. Algorithm~\ref{alg:diffusion_quantization} meticulously delineates the quantization process of an input image, subjected to $T$ diffusion steps, culminating in a quantized representation.

\begin{figure}[!t]
\centering
\includegraphics[width=1\linewidth, keepaspectratio]{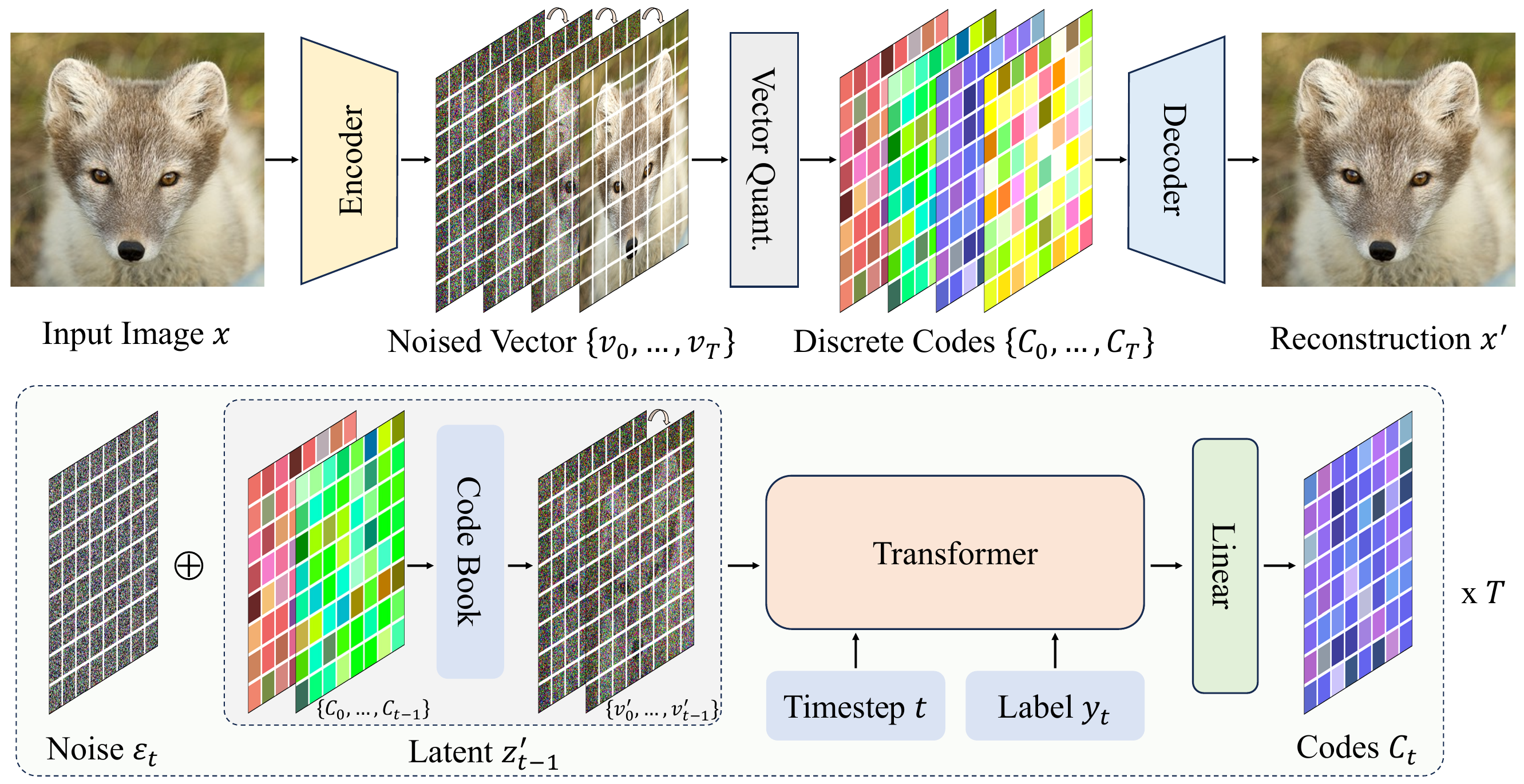}
\caption{
Overview of the RDPM framework. The framework comprises two main stages: 1) Diffusion-based image tokenization, where Gaussian noise is incrementally mixed into the image and quantized to obtain discrete codes, and 2) Recurrent token prediction using a transformer model, which efficiently synthesizes images by recurrently predicting discrete visual codes at each timestep.
}
\label{fig:pip}
\vspace{-5pt}
\end{figure}

\vspace{5pt}
\noindent\textbf{Image Decoding}:
At the final step, \ie, the $T$-th step, we obtain $T$ quantized vectors $\{v'_1, ..., v'_T\}$ and their index codes $\{C_1, ..., C_T\}$, as well as $z'_T = \sum_1^T v'_t$.
The quantized vector representation $z'_T$ is then passed through the decoder $D$ to recover the original image $x'$.
Similar to encoder $E$, the decoder $D$ primarily consists of residual blocks; however, the difference lies in the upsampling of the quantization vector by a multiple of $H/h$.

\vspace{5pt}
\noindent\textbf{Loss Function}:
The loss function is consistent with that used in VQGAN~\cite{yu2021vector}, incorporating several components. The reconstruction loss is defined as:
\begin{equation}
L_{\text{recon}} = \|x - x'\|^2.
\end{equation}
which minimizes the pixel-wise difference between the input image and the reconstructed image. The quantization loss is expressed as:
\begin{equation}
L_{\text{quant}} = \sum_{t=1}^{T} \gamma_t (v_t - v'_t)^2,
\label{eq:gamma}
\end{equation}
representing the weighted average across the $T$ steps to mitigate information loss during vector quantization.
In practice, $\gamma_t = \alpha_t$, as the image's proportion in the noise vector directly influences the model's learning difficulty. The subsequent experimental section further validates the advantageous impact of this weight parameter on image reconstruction performance.
Additionally, the loss calculation for the VAE includes perceptual loss $L_{\text{pept}}$ and adversarial loss $L_{\text{gan}}$:
\begin{equation}
L_{\text{vae}} = L_{\text{recon}} + \delta L_{\text{quant}} + L_{\text{pept}} + \eta L_{\text{gan}}.
\end{equation}
The perceptual loss aligns the high-level semantics of the input and reconstructed images, typically using features extracted from VGG16~\cite{simonyan2014very}, to enhance visual fidelity. Concurrently, the adversarial loss promotes realism through the interaction of a generator and discriminator. The weights $\delta$ and $\eta$ for each loss component are set following VQGAN's configuration, at $0.25$ and $0.75$, respectively.

\begin{algorithm}[!t]
\caption{Diffusion-based Image Quantization}
\label{alg:diffusion_quantization}
\textbf{Input}: Input image $x$, diffusion steps $T$.\\
\textbf{Output}: Discrete codes $\{C_1, \ldots, C_T\}$, noise $\{\epsilon_1, ..., \epsilon_T\}$, accumulated representation $z'_T$.

\begin{algorithmic}[1]
\STATE \textbf{Initialize:} $v = E(x)$, $z_0 = v$, $v'_0 = 0$, $z'_0 = 0$.
\FOR{$t = 1$ to $T$} 
    \STATE $z_t = z_{t-1} - v'_{t-1}$, update the residual $z_t$.
    \STATE $v_t = \alpha_t z_t + \beta_t \epsilon_t$, mix Gaussian noise.
    \STATE $v_t' = Q(v_t)$, quantize $v_t$ and obtain $C_t$.
    \STATE $z'_t = z'_{t-1} + v'_t$, accumulate the quantized vector.
    % \STATE $C_t = \arg \min_{c \in \mathcal{C}} | v_{t} - e(c) |_2^2$, nearest neighbor search.
\ENDFOR
\RETURN $\{C_1, ..., C_T\}$, $\{\epsilon_1, ..., \epsilon_T\}$, \text{and } $z'_T$.
\end{algorithmic}
\end{algorithm}

\subsection{Recurrent Token Prediction for Generation}
The goal of a generative model is to learn how to transition from random noise to a target data distribution within the latent space over $T$ iterative steps.

\vspace{5pt}
\noindent\textbf{Training Process}:
The input image $x$ is encoded to yield $T$ quantized vectors of size $h \times w \times d'$, denoted as $\{v'_1, ..., v'_T\}$, and corresponding codes of size $h \times w$, $\{C_1, ..., C_T\}$, alongside the accumulated noise $\{\epsilon_1, ..., \epsilon_T\}$ from the diffusion process.
At each step $t$, all known factors are input into a generative model $f$. These factors include the noise $\epsilon_t$, a condition (\ie, category label $y$), the time step $t$, and the accumulated quantized vectors from step $1$ to $t-1$, $z'_{t-1}$, to predict $v'_t$ (with $z'_{t-1} = 0$ when $t = 1$):
\begin{equation}
\widehat{v}'_t = f(\epsilon_t, y, t, z'_{t-1}).
\end{equation}
The objective is to minimize the discrepancy between the predicted $\widehat{v}'_t$ and the ground truth $v'_t$.
In practice, noise and quantized vectors are concatenated to form a variable of size $h \times w \times 2d$, which is then projected back to $h \times w \times d$.
After unfolding into $h \times w$ sequence tokens and processing through a transformer model, a classification layer predicts the quantization code index $C_t$ at step $t$, using cross-entropy loss:
\begin{equation}
L_{gen} = -\frac{1}{h \times w} \sum_{i=1}^{h \times w} \sum_{j=1}^{K} C_{t,i,j} \log\left(\frac{e^{\widehat{v}'_{t,i,j}}}{\sum_{k=1}^{K} e^{\widehat{v}'_{t,i,k}}}\right).
\end{equation}
Here, $i$ indexes the token position, while $j$ and $k$ index the embedding positions in the codebook. Algorithm~\ref{alg:image_generation} succinctly encapsulates the predictive procedure, beginning with stochastic noise and sequentially predicting the subsequent data distribution over  $T$ iterations in a recurrent manner. The aggregated distribution is then input into the decoder, culminating in the generation of the desired image.

\begin{algorithm}[!t]
\caption{Recurrent Token Prediction}
\label{alg:image_generation}
\textbf{Input}: Class labels $y$, decoder $D$.\\
\textbf{Output}: Generated image $\widehat{x}'$.

\begin{algorithmic}[1]
\STATE \textbf{Initialize:} $\widehat{z}'_0 = 0$.
\FOR{$t = 1$ to $T$} 
    \STATE $\widehat{v}'_t = f(\epsilon_t, y, t, \widehat{z}'_{t-1})$, predict the direction of the next data distribution $v'_t$, where $\epsilon_t \sim \mathcal{N}(0, 1)$.
    % \STATE $\widehat{C}_t = \text{argmax}(\widehat{v}'_t)$, calculate the predicted code.
    \STATE $\widehat{z}'_t = \widehat{z}'_{t-1} + \widehat{v}'_t$, update the accumulated distribution estimate.
\ENDFOR
\STATE $\widehat{x}' = D(\widehat{z}'_T)$, decode the predicted data distribution.
\RETURN $\widehat{x}'$.
\end{algorithmic}
\end{algorithm}

\vspace{5pt}
\noindent\textbf{Testing Phase}:
During the testing phase, the prediction of $v'_t$ is conducted based on known conditions, following the formula $\widehat{z}'_t = \widehat{z}'_{t-1} + f(\epsilon_t, y, t, \widehat{z}'_{t-1})$. Here,  when $t=0$, $\widehat{z}'_{t-1} = 0$.
This recurrent forward propagation over $T$ steps yields the data distribution $\widehat{z}'_T$.
Subsequently, this distribution is processed through the decoder to generate the final image. The incorporation of random noise at each step of the $T$-step prediction process ensures the diversity of the generated outputs.

In addition, we employ the widely adopted Classifier-Free Guidance (CFG) strategy~\cite{ho2022classifier}. During training, the condition 
y
 is occasionally set to empty with a certain probability to facilitate learning. In the inference stage, conditional prediction is refined by subtracting the unconditional prediction from the conditional prediction, expressed as $\widehat{v}'_t = \text{cond}(\widehat{v}'_t) + (\text{cond}(\widehat{v}'_t) - \text{uncond}(\widehat{v}'_t)) \lambda$, where $\lambda$ denotes the strength of guidance.
Furthermore, inspired by MaskGIT~\cite{chang2022maskgit}, we incorporate Gumbel noise~\cite{jang2016categorical} to enhance the prediction of the code \( C_t \):
\begin{equation}
\widehat{C}_t = \text{argmax}(\widehat{v}'_t \tau -\log(-\log(u))),
\end{equation}
where $u \sim \text{Uniform}(0, 1)$ is noise sampled from a uniform distribution over the interval $[0-1]$, and $\tau$ is the temperature parameter.
This approach further augments the diversity of the RDPM in the image generation process.

\section{Experiment}

In this section, we provide a comprehensive overview of our experimental setup, detailing the datasets, evaluation metrics, and parameter configurations employed in our study. We perform ablation studies to scrutinize the impact of hyperparameter selection. Furthermore, we conduct a thorough comparison of our method against SOTA models.

\begin{table}[!t]
\centering
\small
\caption{
Specifications of diffusion models across various sizes.
}
\label{tbl:abl_param_set}
\renewcommand{\arraystretch}{1.0}
\begin{tabular}{ccc|c}
\toprule
depth  &  hidden size & head num   & Params.  \\
\midrule
12 & 768 & 12 & 133M \\
16 & 1,024 & 16 & 310M \\
20 & 1,280 & 20 & 602M \\
\bottomrule
\end{tabular}
\vspace{-5pt}
\end{table}

\subsection{Datasets and Evaluation Metrics}
Our experiments involve training image autoencoder and generative models using the ImageNet-1k training dataset~\cite{deng2009imagenet}, which consists of 1.28 million images.

To assess the quality of the generated images, we primarily employ the Fréchet Inception Distance (FID). This metric quantifies the divergence between the distributions of generated images and reference images by comparing their mean and covariance matrices. A lower FID score indicates a closer alignment between the distributions, suggesting higher image quality. Additionally, we use the Inception Score (IS) to evaluate the diversity and quality of the generated images, based on the entropy of their classification probability distribution. Precision and Recall metrics are also utilized to illustrate the overlap between the distributions of generated and reference images, where higher Precision signifies better image quality and higher Recall indicates greater diversity.

For evaluating the reconstruction performance of the VAE model, we use variants of the FID, specifically the reconstruction FID (rFID), along with the Peak Signal-to-Noise Ratio (PSNR) and the Structural Similarity Index (SSIM). These metrics assess distributional differences, pixel-level fidelity, and structural similarity between the original and reconstructed images, respectively. In calculating the FID score, we generate 50 images for each of the 1,000 categories, resulting in a total of 50K images.

\begin{table}[!t]
\centering
\small
\caption{
Performance of VAE under different settings.
The term `weight' refers to the activation of the $\gamma_t$ coefficient in Eq.~\ref{eq:gamma}, while `bias' denotes the addition of a shared convolutional layer to process quantized vectors as bias at each timestep.
}
\label{tbl:abl_vae}
\renewcommand{\arraystretch}{0.95}
\resizebox{0.47\textwidth}{!}{
    \begin{tabular}{cccc|ccc}
    \toprule
    Exp. & weight  & bias   & $T$ & rFID ($\downarrow$) & SSIM ($\uparrow$) & PSNR ($\uparrow$)  \\
    \midrule
     1 &  -    &    -    & 6  & 3.58 & 0.4890 & 29.45 \\
     2 &  -    &    -    & 8  & 3.21 & 0.5133 & 29.46 \\
     3 &  -    &    -    & 10 & 3.05 & 0.5657 & 29.56 \\
     4 &  -    &    -    & 12 & 3.01 & 0.5669 & 29.60 \\
     5 &  -    &    -    & 14 & 2.93 & 0.5761 & 29.61 \\
     6 &  -    &    -    & 20 & 2.90 & 0.5790 & 29.64 \\
     7 & $\surd$ &    -    & 10 & 2.04 & 0.6178 & 29.71 \\
     8 &  -    & $\surd$ & 10 & 2.06 & 0.5994 & 29.89 \\
     9 & $\surd$ & $\surd$ & 10 & 1.52 & 0.6226 & 29.97 \\
    \bottomrule
    \end{tabular}
}
\vspace{-10pt}
\end{table}

\subsection{Experimental Setting}
In the training phase of the image tokenization module, input images with dimensions of $256 \times 256$ are downscaled by a factor of 16, resulting in $16 \times 16$ tokens. The codebook utilized consists of $4096$ vectors, each with an embedding dimension of 32. We employ the Adam optimizer with betas $[0.5, 0.9]$, setting the weight decay to $0$. The learning rate is maintained at $4e^{-5}$, with a warmup period spanning 0.5 epochs. The batch size is configured to 128, and training is executed over 10 epochs using the ImageNet-1k dataset. Compared to VQ-GAN~\cite{esser2021taming}, our model incorporates a minimal increase in parameters, including the addition of a Layer Normalization~\cite{ba2016layer} layer and a convolutional layer. Specifically, the encoder's latent vector undergoes layer normalization to harmonize its scale with noise prior to diffusion-based quantization. Furthermore, a shared convolutional layer is introduced during the quantization phase, serving as a bias for the quantized vectors, which enhances image reconstruction as substantiated by subsequent experimental results.

For the generative model training, we follow the LLama~\cite{touvron2023llama} framework, setting the transformer's hidden size to $depth \times 64$, with the number of self-attention heads equaling the depth. Table~\ref{tbl:abl_param_set} provides detailed specifications. The probability of randomly nullifying the category condition is $10\%$. The time-step $t$ and category condition $y$ are integrated using AdaLN-Zero~\cite{peebles2023scalable}. To expedite training, mixed-precision techniques are employed, and flash-attention v2~\cite{dao2023flashattention} is utilized to enhance the efficiency of attention operations and reduce memory usage. The batch size is set to $1024$, with an initial learning rate of $4e^{-4}$, and a warmup period of $1000$ steps. Training extends over 200 to 250 epochs, with gradient clipping set to 1. The AdamW optimizer is applied with betas $[0.9, 0.95]$ and a weight decay of $0.03$. Model parameters are updated using an exponential moving average (EMA) with a decay rate of  $0.9999$.

% \begin{figure*}[!t]
% \centering
% \includegraphics[width=0.9\linewidth, keepaspectratio]{tmp.pdf}
% \caption{
% Reconstruction and generation results on various parameter settings of xxx.
% }
% \label{fig:vae}
% \end{figure*}

\subsection{Ablation Study}
This subsection presents an extensive experiments aimed at validating the selection of hyperparameters in training processes of both image autoencoder and generative models.

As depicted in Table~\ref{tbl:abl_vae}, Exp.1-6, we investigate the impact of varying the timestep $T$ on an image tokenization model based on the diffusion process.
When $T$ is less than $10$, the reconstruction performance significantly deteriorates. Conversely, increasing $T$ beyond $10$ results in only marginal improvements across various metrics.
Consequently, we have fixed $T$ at $10$.
Upon further analysis, we observe that latent quantization loss increases as the timestep decreases and the noise mixing ratio rises. Additionally, images reconstructed in the initial timesteps are heavily infused with noise. Given the codebook's limited capacity (\ie, $4096$),  the extent of random noise mixing directly influences information loss during quantization and the model's learning difficulty. 
To address this, as shown in Eq.~\ref{eq:gamma}, we introduce a weight $\gamma_t$ at each timestep, proportional to the image latent vector $\alpha_t$.
This adjustment reduces the rFID to $2.04$, as indicated Exp.7 in Table~\ref{tbl:abl_vae}. 
Moreover, incorporating an additional timestep-shared convolutional layer to process quantized vectors and using these vectors as bias can further mitigate quantization loss at smaller timesteps, achieving an rFID of $2.06$ (Exp. 8).
By combining both strategies, the rFID decreases to $1.52$, with improvements observed in SSIM and PSNR metrics.

\begin{table}[t]
\centering
\small
\caption{
Various sampling strategies in the image synthesis model with depth of $12$.
}
\label{tbl:abl_gen}
\renewcommand{\arraystretch}{0.95}
\resizebox{0.47\textwidth}{!}{
    \begin{tabular}{l|ccc>{\centering\arraybackslash}m{10.5mm}c}
    \toprule
    Strategy & FID-50K   & sFID     & IS  & Precision  & Recall \\
    \midrule
    ~~~Raw    & 19.38& 13.66 & 69.48 & 0.62 & 0.58\\
    % +~RoPE\\
    +~CFG     & 6.59 & 13.93 & 215.5 & 0.84 & 0.41\\
    +~CFG linear & 5.31 & 11.90 & 233.2 & 0.84 & 0.45\\
    +~Gumbel  & 3.68 & 6.30  & 244.6 & 0.84 & 0.49 \\
    \bottomrule
    \end{tabular}%
}
%\vspace{-2mm}
\end{table}%
\begin{table}[!t]
\centering
\small
\caption{
Variance schedule of gradually adding Gaussian noise to the latent vector.
The experiment is conducted based on a model with a depth of $12$ and trained for $100$ epochs.
}
\label{tbl:abl_schedule}
\renewcommand{\arraystretch}{0.95}
\begin{tabular}{l|ccccc}
\toprule
Schedule & FID-50K   & sFID     & IS  & Precision  & Recall \\
\midrule
$sin$  & 5.75 & 11.90 & 246.4 & 0.86 & 0.39 \\
$linear$ & 4.94 & 7.12 & 243.0 & 0.86 & 0.42 \\
$pow$  & 4.30 & 6.29 & 221.9 & 0.84 & 0.47 \\
\bottomrule
\end{tabular}%
%\vspace{-5pt}
\end{table}%

% %
% \begin{figure}[!t]
% \centering
% \includegraphics[width=1\linewidth, keepaspectratio]{./variance_schedule.pdf}
% \caption{
% %
% Images generated by different sampling methods from the depth-$12$ model.
% %
% }
% \label{fig:abl_vis}
% \end{figure}
% %

During the inference phase of our image synthesis model, as illustrated in Table~\ref{tbl:abl_gen}, we assess the benefits of various sampling strategies. Initially, applying CFG~\cite{ho2022classifier} results in substantial accuracy improvements for generated images, albeit at the cost of reduced recall. This suggests that the CFG strategy enhances class semantic information but compromises diversity. Building on the MaskGit~\cite{chang2022maskgit} approach, we modify the constant guidance value at each step to a linearly increasing process. Specifically, we employ a lower guidance value in the initial timesteps to preserve diversity and gradually intensify class guidance in later steps. As shown in the third row of Table~\ref{tbl:abl_gen}, this approach maintains accuracy while enhancing recall (\ie, diversity). Additionally, inspired by MaskGit, we introduce Gumbel noise into predictions during inference, further boosting the diversity of synthesized images.

\begin{figure}[!t]
\centering
\includegraphics[width=1\linewidth, keepaspectratio]{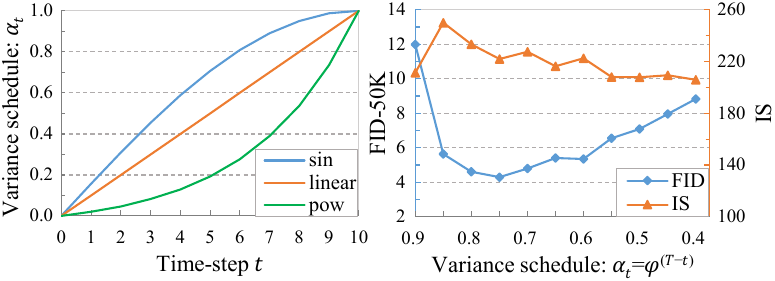}
\caption{
(Left) Variation of $\alpha_t$ across different scheduling strategies. (Right) Changes in FID as the hyperparameter $\varphi$ varies under the $pow$ schedule.
The experiment is conducted on the model with depth of $12$ and trained for $100$ epochs.
}
\vspace{-5pt}
\label{fig:var_sch}
\vspace{-10pt}
\end{figure}

In the process of incrementally integrating Gaussian noise into the latent vector, as depicted in Figure~\ref{fig:var_sch}, we evaluate three variance schedules: $sin$, $linear$, and $pow$, corresponding to $\alpha_t = \sin(\pi/2 * t/T)$, $\alpha_t = t/T$, and $\alpha_t = \varphi^{(T-t)}$, respectively. 
These schedules represent different progression trends for $\alpha_t$ from 0 to 1.
As shown in Table~\ref{tbl:abl_schedule}, the $pow$ schedule achieves the best FID-50K result for the generative model. Although IS and Precision are not optimal, the generated images' distribution aligns most closely with the reference images. Therefore, we adopt the $pow$ schedule in subsequent experiments. Furthermore, since $\varphi$ is adjustable, we further evaluate the $pow$ schedule's performance with various $\varphi$ settings in Figure~\ref{fig:var_sch}.
When $\varphi$ equals $0.75$, the generative model achieves the best FID-50K result. Although the IS performance is not the best in this setting, we prioritize the FID metric to balance the quality and diversity of generated images.

\begin{table*}[!t]
\centering
\small
\caption{
Performance comparison on the $256 \times 256$ ImageNet class-conditional image generation task.
The model with `-re' suffix indicates employing the rejection sampling strategy.
}
\label{tbl:sota}
\renewcommand{\arraystretch}{0.95}
\begin{tabular}{c|l|l|cc|cccc}
\toprule
Type & Tokenizer & Model   & Params.     & Steps  & FID~$\downarrow$  & IS~$\uparrow$ & Precision~$\uparrow$ & Recall~$\uparrow$ \\
\midrule
GAN & - & BigGAN~(Brock~et~al.~2018) & 112M & 1 & 6.95 & 224.5 & 0.89 & 0.38 \\
GAN & - & GigaGAN~(Kang~et~al.~2023) & 569M & 1 & 3.45 & 225.5 & 0.84 & 0.61 \\
GAN & - & StyleGAN-XL (Sauer~et~al.~2022) & 166M & 1 & 2.30 & 265.1 & 0.78 & 0.53 \\
\midrule
Continuous Diff. & VAE & ADM~(Dhariwal~et~al.~2021) & 554M & 250 & 10.94 & 101.0 & 0.69 & 0.63 \\
Continuous Diff. & VAE & CDM~(Ho~et~al.~2021) & - & 8,100 & 4.88 & 158.7 & - & - \\
Continuous Diff. & VAE & LDM-4~(Rombach~et~al.~2022) & 400M & 250 & 3.60 & 247.7 & - & - \\
Continuous Diff. & VAE & GIVT~(Michael~et~al.~2023) & 304M & 256 & 3.35 & - & 0.84 & 0.53 \\
Continuous Diff. & VAE & DiT-XL/2~(Peebles~et~al.~2023) & 675M & 250 & 2.27 & 278.2 & 0.83 & 0.57 \\
\midrule
Continuous AR & VAE & MAR~(Li~et~al.~2024) & 208M & 64 & 2.31 & 281.7 & 0.82 & 0.57 \\
Continuous AR & VAE & MAR~(Li~et~al.~2024) & 479M & 64 & 1.78 & 296.0 & 0.81 & 0.60 \\
Continuous AR & VAE & MAR~(Li~et~al.~2024) & 943M & 64 & 1.55 & 303.7 & 0.81 & 0.62 \\
\midrule
Discrete AR & VQ-VAE & VQGAN~(Esser~et~al.~2021) & 227M & 256 & 18.65 & 80.4 & 0.78 & 0.26 \\
Discrete AR & VQ-VAE & VQGAN~(Esser~et~al.~2021) & 1.4B & 256 & 15.78 & 74.3 & - & - \\
Discrete AR & VQ-VAE & VQGAN-re~(Esser~et~al.~2021) & 1.4B & 256 & 5.20 & 280.3 & - & - \\
Discrete AR & VQ-VAE & ViT-VQGAN~(Yu~et~al.~2021) & 1.7B & 1,024 & 4.17 & 175.1 & - & - \\
Discrete AR & VQ-VAE & ViT-VQGAN-re~(Yu~et~al.~2021) & 1.7B & 1,024 & 3.04 & 227.4 & - & - \\
Discrete AR & VQ-VAE & RQTran.~(Lee~et~al.~2022) & 3.8B & 68 & 7.55 & 134.0 & - & - \\
Discrete AR & VQ-VAE & RQTran.-re~(Lee~et~al.~2022) & 3.8B & 68 & 3.80 & 323.7 & - & - \\
% \midrule
Discrete AR & VQ-VAE & MaskGIT~(Chang~et~al.~2022) & 227M & 8 & 6.18 & 182.1 & 0.80 & 0.51 \\
Discrete AR & VQ-VAE & MaskGIT-re~(Chang~et~al.~2022) & 227M & 8 & 4.02 & 355.6 & - & - \\
Discrete AR & VQ-VAE & VAR-d16~(Tian~et~at.~2024) & 310M & 10 & 3.30 & 274.4 & 0.84 & 0.51 \\
Discrete AR & VQ-VAE & VAR-d20~(Tian~et~at.~2024) & 600M & 10 & 2.57 & 302.6 & 0.83 & 0.56 \\
\midrule
Discrete Diff. & VQ-VAE & RDPM~(depth=12) & 133M & 10 & 3.68 & 244.6 & 0.84 & 0.49 \\
Discrete Diff. & VQ-VAE & RDPM~(depth=16) & 310M & 10 & 2.91 & 270.3 & 0.83 & 0.54 \\
Discrete Diff. & VQ-VAE & RDPM~(depth=20) & 602M & 10 & 2.56 & 295.1 & 0.80 & 0.58 \\
\bottomrule
\end{tabular}%
\vspace{-10pt}
\end{table*}%

\subsection{Comparison with SOTA methods}
We next evaluate the performance of various generative models for class-conditional image generation at the $256 \times 256$ ImageNet scale. Table~\ref{tbl:sota} summarizes the results of all state-of-the-art models.

From the table, it is clear that GAN-based methods, such as StyleGAN-XL~\cite{sauerscaling}, achieve outstanding FID scores, underscoring their effectiveness. The one-step sampling process of these models further highlights their efficiency. However, they encounter challenges related to training stability, which may hinder their scalability to more complex tasks, such as text-to-image generation.

Diffusion models simplify the generation process by decomposing the data distribution transformation into $1000$ steps, thereby significantly reducing the complexity of image generation. Coupled with a robust VAE from StableDiffusion~\cite{rombach2022high}, DiT-XL/2~\cite{peebles2023scalable} achieves a FID of $2.27$, with balanced Precision and Recall indicating a closer alignment with the reference image distribution. While diffusion-based methods are hindered by the number of sampling steps (\eg, 250), acceleration strategies like iDDPM~\cite{nichol2021improved} and DPM-solver~\cite{lu2022dpm} can reduce this to tens of steps, albeit at the cost of image quality.

Discrete autoregressive methods, such as VQGAN~\cite{esser2021taming} and ViT-VQGAN~\cite{yu2021vector}, rely on predicting the raster scan order and often exhibit inefficiency during the sampling process, typically requiring numerous iterations to achieve satisfactory results. However, their compatibility with large language models (LLMs) offers substantial scaling potential. For example, RQTran.~\cite{lee2022autoregressive} scales up to 3.8B parameters while reducing sampling steps to $68$. Additionally, it is well-known that VQ-VAE~\cite{razavi2019generating} utilizes discrete representations, leading to information loss during quantization, which results in inferior performance compared to other methods. Mask-based image synthesis strategies achieve moderate FID scores but excel in sampling efficiency, requiring just eight steps. In this domain, VARVAR~\cite{tian2024visual} distinctively employs a coarse-to-fine autoregressive (AR) generation strategy to attain superior performance.

Our proposed paradigm strikes a superior balance between sampling efficiency and image quality. Specifically, RDPM delivers high-quality image generation with only 
10 timesteps and fewer parameters than autoregressive models. With 133M parameters, RDPM surpasses all autoregressive models without the need for rejection sampling~\cite{razavi2019generating,esser2021taming,chang2022maskgit}. Even at the same parameter scale, RDPM matches diffusion-based methods in quality, with a notable advantage in model forward passes. Besides, the results indicate that our approach achieves performance on par with VAR. It is crucial to emphasize that our primary objective is not to achieve absolute SOTA performance. Instead, our focus is on introducing a novel paradigm for image generation, specifically through the lens of discrete diffusion.

\subsection{Discussion}

MAR~\cite{li2024autoregressive} has skillfully combined masked token prediction techniques with diffusion methods, effectively modeling the per-token probability distribution through a diffusion process. Concurrently, other research has explored various fusion strategies for integrating text autoregressive models with image diffusion techniques, thereby unifying text and visual generation tasks, as exemplified by ShowO~\cite{xie2024show} and Transfusion~\cite{zhou2024transfusion}. Another sophisticated model, VAR~\cite{tian2024visual}, employs a progressive generation technique based on resolution. However, this approach is confined to the realm of image generation and can not be generalized to other continuous signal domains, such as speech synthesis, where the concept of resolution is absent.

In contrast to these methodologies, our work introduces a novel perspective by implementing a diffusion process through multi-step discrete token prediction, aligning with the next token prediction strategy commonly employed in large language models (LLMs). This approach not only maintains the optimization consistency inherent in LLMs' next token prediction but also addresses the limitations associated with image quality loss that arise from single-step prediction mechanisms. Consequently, this method fills a critical gap in the field of Discrete Diffusion models, as illustrated in Fig~\ref{fig:taxonomy}, and paves the way for the integration of discrete and continuous signal generation models.
\section{Conclusion}

In conclusion, we introduce a novel paradigm for image generation, termed RDPM, which gradually denoises and ultimately generates natural images from a standard Gaussian distribution by recursively predicting the discrete codes for subsequent timesteps. This work addresses a significant gap in the field of Discrete Diffusion within image generation. Within RDPM, we propose a diffusion-based image tokenization method that effectively converts images into a sequence of timestep-based discrete codes, a technique that is also applicable to other continuous signal domains. Our approach has demonstrated impressive results on the ImageNet dataset, achieving state-of-the-art performance among methods employing discrete tokenizers. We anticipate that this method will inspire new directions in the development of unified multimodal models.
% \clearpage
{
    \small
    \bibliographystyle{ieeenat_fullname}
    \bibliography{main}
}

% WARNING: do not forget to delete the supplementary pages from your submission 
% \input{sec/X_suppl}

\end{document}